# Analyzing Machine Learning Models for Credit Scoring with Explainable AI and Optimizing Investment Decisions

Swati Tyagi
Institute of Financial Service Analytics
Alfred Lerner College of Business and Economics
University Of Delaware

**Abstract:** This paper examines two different yet related questions related to explainable AI (XAI) practices. Machine learning (ML) is increasingly important in financial services, such as pre-approval, credit underwriting, investments, and various front-end and back-end activities. Machine Learning can automatically detect non-linearities and interactions in training data, facilitating faster and more accurate credit decisions. However, machine learning models are opaque and hard to explain, which are critical elements needed for establishing a reliable technology. The study compares various machine learning models, including single classifiers (logistic regression, decision trees, LDA, QDA), heterogeneous ensembles (AdaBoost, Random Forest), and sequential neural networks. The results indicate that ensemble classifiers and neural networks outperform. In addition, two advanced post-hoc model agnostic explainability techniques - LIME and SHAP are utilized to assess ML-based credit scoring models using the open-access datasets offered by US-based P2P Lending Platform, Lending Club. For this study, we are also using machine learning algorithms to develop new investment models and explore portfolio strategies that can maximize profitability while minimizing risk.

## I.    Introduction

Credit risk analysis has gained increasing importance since the sub-prime mortgage crisis of 2007 and the global financial crisis of 2008. Financial institutions use credit scoring systems to assess their credit risks. An automated credit approval process can gather all the necessary information, evaluate the loan application, and decide if it is approved. Credit scores can be created by analyzing available data using statistical and machine learning techniques so that this process can be automated. In some cases, these models lack transparency, leaving model developers to guess what the models might have learned. Ribeiro et al. (2016) "pointed out that if machine learning classifiers are used directly as decision-making tools, a significant concern arises: if the users do not trust the model's predictions, they will not use it."

Regulations also require explanations for credit decisions. Financial service providers in Europe face this issue of explainability because of the General Data Protection Regulation (GDPR) that applies throughout the entire European Union. A principal benefit of the GDPR is that it provides people with meaningful information about automated decisions which directly affect them if they are solely based on automated processing. In addition, the US Equal Housing Lender rule has provided another reason to focus on model explainability. Therefore, describing a model's output appears important for facilitating innovation in financial applications. As a new generation of AI models, eXplainable AI (XAI) employs machine learning (ML) to develop models that balance explainability with predictive fitness, enabling humans to operate and trust AI. In explaining how machine learning works, these two frameworks have been widely respected:

- "The LIME framework", introduced by Ribeiro et al. (2016)
- "SHAP values", introduced by Lundberg et al. (2017)

Selecting loans that have a high chance of success becomes critical when creating a portfolio with a low risk and high return. There have been some studies on credit risk in peer-to-peer lending, assessing how borrowers behave and predicting credit risk to help lenders select loans. Credit scoring has become the primary method used for evaluating borrowers' risks in recent years. Few studies have incorporated profit-scoring approaches into their risk-prediction algorithms for P2P loans. This study uses an approach that analyzes risk and return across the entire portfolio to select loans with low risk. A previously unexplored approach will be used in this study: Our research worked towards validating the train/test split approach and compared why standardized K-fold cross validation is a better approach when class imbalance is present. Finally, we show how defining the target function correctly is integral to optimizing the highest payout.

In section 2, we review the literature. In Section 3, the methodology is explained in detail, and in Section4, the results are discussed. We will discuss the modeling procedure in section 5 and analyze the comparative results of several machine models applied to loan default prediction, including the SHAP and LIME for neural network models, as well as other techniques such as model-based profit maximization and portfolio-based profit maximization using expected returns. Section 6 consists of some conclusions and recommendations for future research.

## II. Literature Review

As part of this section, we present the existing work related to the field of credit scoring for lending purposes. A growing body of literature explores various methods of predicting default. There are many empirical studies conducted to date, ranging from applications of "univariate and bivariate models", (e.g., Serrano-Cinca et al. (2015)) to applying advanced analytics, such as machine learning (for example, Malekipirbazari et al. (2015), deep learning (for example, Bastani et al. (2019), and neural networks (for example, Babaev et al.(2019)). Kroppa (2013), "makes a case for the use of machine learning methods such as Random Forests (RF) to estimate individual credit risk". Aniceto et al.(2020)'s study proves that RF outperforms industry standard logistic regression. According to Khandani et al. (2010), "bootstrapped CART trees outperform industry-standard models in predicting rates of credit-card defaults and delinquencies". Different studies have analyzed various approaches for predicting loan defaults and assessing credit worthiness. Numerous studies use the Lending Club dataset as fundamental raw data for machine learning.

**Machine Learning Explainability** A study by Bussman et al. (2020) uses Shapley values to construct an XAI model for fintech risk management in which the authors use a sample of 15,000 Small Medium Enterprise (SME) firms to explain loan decisions. However, the authors do not consider some of the main technical challenges of using Shapley values in larger data sets, which would be more appropriate for financial service providers. A study of Shapley values has shown that they help improve the transparency of complex ML models applied to credit risk. The most exciting precedent may be the work by Miller et al. (2020), where they assessed the predictive ability of well-known ML models such as Random Forests, SVM etc. for the scoring of P2P lending platforms' credit and also explaining them using SHAP.

## III. Methodology

### 3.1 Description

Data from the Lending Club website was used for this study(2016year). It contained110 variables in its raw format. During the process of cleaning and interpreting the dataset, several transformations were performed, including removing the variable with more than 30% missing values. In total, 58 variables were removed. Variables such as URL, member, and ID have been removed. In the end, there were approximately 110,000 records in a new feature space with 30 features. To handle percentages, dates, and numeric factors, we performed several transformations. For this purpose, we used variables which were only available from the credit application. Factor variables that represent an event or condition after funding the credit are excluded. Credit states canbe1(Paid100%) or 0 (Default). Our selection process is limited to accounts completed within the everyday creditlife-36or60months.Wehaveappliedanoutliertreatmenttoensurethedataset's quality. Several variables have been treated with appropriate methods such as outlier removal, capping, and discretization, depending on the percentages and distributions of the outliers. For this study, models are divided into three groups:

1. Individual classifiers– neural networks, logistic regression, LDA, QDA.
2. Homogeneous ensemble classifiers – combining several models based on the same algorithm. These models have in common that they combine several weak classifiers into one strong.
3. Heterogeneous ensemble classifiers - unlike homogeneous ensembles, are included in models based on different algorithms.

We then further purified the input space by removing highly correlated features. These features were identified by plotting heat maps with correlations between numerical features. Chi-square test was used to select categorical variables. All categories were encoded using one hot coding. A binary categorical variable was created for the target variable, loan status("Status"). We retained only two categories, "Fully Paid" and "Default". Our credit risk management process included the following steps:

**Assessing Collinearity**: It seems that we are having a problem with Collinearity for the revolving columns such as revol_util, revol_balance et al. We removed features that has a correlation more than 0.8 with any other feature.

The dataset used to train the model cannot be used to determine the model's accuracy. Subsets of the data used to train the model, while others held back to analyzed trained models on test data. Creating a model is a big problem because the point is to make predictions based on new data. Therefore, to estimate accuracy we also used resampling method (stratified 10-Fold cross validation) to resample the training dataset.

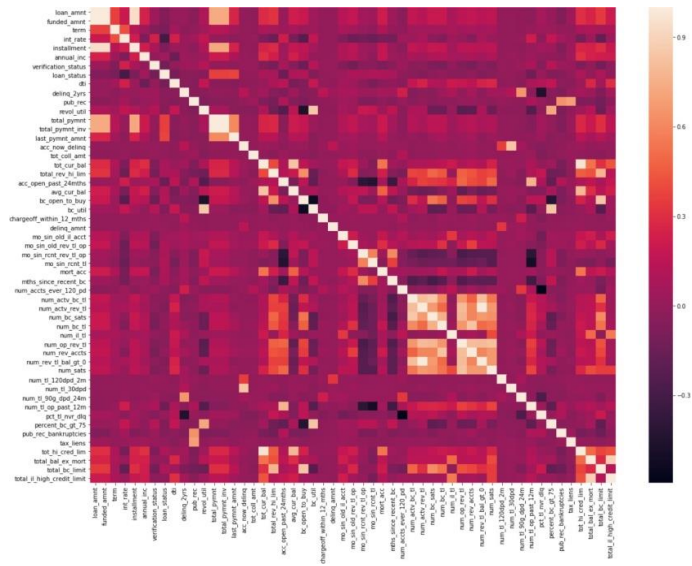

Figure1: Assessing Co-linearity of Common features

- Create training and test sets from a dataset.
- Estimate the accuracy of algorithms using stratified k-fold(10-fold) cross-validation.

## IV. Algorithms used for Credit Risk Management

**Logistic Regression** Logistic regression is a binary classification method that is most used for analyzing datasets that are subjected to one or more independent variables that influence the outcome. As a discrete binary output, the output from this function is a probability between 0 and 1, and the function is used to model the conditional probabilities $P(Y=1|X=x)$ and $P(Y=0|X=x)$. It is based on the linear regression model, using the sigmoid function to compress the result of the linear model $w^tx$ to [0,1]. As we all know, linear regression can be described as the following equation:

$$w_x = w_0 + w_1x_1 + w_2x_2 + w_nx_n \quad (1)$$

Formally, the logistic regression model is as follows:

$$p(Y=1|X=x) = \frac{exp(wx+b)}{1+exp(wx+b)} \quad (2)$$

$$p(Y=0|X=x) = \frac{1}{1+exp(wx+b)} \quad (3)$$

And the sigmoid function is:

$$y = \frac{1}{1 + exp(-x)}$$

(4)

Logistic regression generates a discrete binary result between 0 and 1, indicating the probability that a sample belongs to one of the categories discussed above.

**Decision Tree** It mimics how humans think with a flowchart structure. This algorithm uses the tree representation to make classification choices:

1. The original set S is used as the root
2. As each algorithm iteration is repeated, the algorithm calculates the very unused attribute's Entropy(H) and Information Gain (IG)
3. After that, it selects the attribute that has the lowest entropy of the most significant information gain
4. A subset of the data is then created from the selected attribute
5. On each subset, the algorithm recurs, considering only attributes never previously considered

In sklearn, the Decision Tree Classifier can give probabilities but must use max depth to truncate the tree. This probability is $P = n_A/(n_A + n_B)$, i.e., the total number of observations of class A that that leaf has captured over the entire number of observations captured by that leaf. In spite of that, we must prune or truncate the decision tree because otherwise, it grows until n=1 in each leaf, thereby equaling P=1.

$$E(S) = \sum_{i=1}^{c} -P_i log_2 P_i$$

(5)

**LDA & QDA** A linear subspace can be of the directions that maximize the separation of classes by using linear discriminant analysis (LDA) to reduce the supervised dimensionality of the input data. In general, output dimensions are smaller than classes, affecting dimensionality reduction significantly. However, this only applies to multiclass problems. A linear combination of variables that most effectively explains data is the goal of both LDA and PCA. Specifically, class differences are taken into account with LDA. Unlike PCA, factor analysis builds feature combinations based on differences instead of similarities. The discriminant analysis technique differs from factor analysis in that it is not based on interdependence: it requires a distinction between independent and dependent variables (also known as criterion variables). We can derive LDA and QDA using simple probability models, which model the class conditional distribution of the data, $P(X|y=k)$ for each class. Bayes' rule can then be used to derive predictions., for each training sample $x \in R^d$:

$$p(Y=k|x) = \frac{P(x|y=k)P(y=k)}{P(x)} = \frac{P(x|y=k)P(y=k)}{\sum_l P(x|y=l).P(y=l)}$$

(6)

QDA is not that much different from LDA except that we assume that the covariance matrix can be different for each class, and so, we will estimate the covariance matrix $\Sigma_k$ separately for each class k, k = 1, 2,…..K.

Quadratic discriminant function:

$$\delta_k(x) = -\tfrac{1}{2}\log|\Sigma_k| - \tfrac{1}{2}(x-\mu_k)^T \Sigma_k^{-1}(x-\mu_k) + \log \pi_k$$

(7)

**Ensemble Methods**

**Random Forest** The Random Forest creates multiple decision trees in the forest. Decision trees that have a supervised performance on classification and regression are widely used in economics due to their robustness and precision. Nonlinear data can also be worked with Random Forest, unlike linear models. A formal model would look like this:

$$G(x) = f_0(x) + f_1(x) + …. + f_n(x)$$

(8)

Given a training part, X=x1…,xn with responses Y=y1…,yn, random samples are selected (B times) with replacement to train decision trees then, the prediction for unseen sample x' is obtained by averaging the predictions from all the trained individual decision trees on x':

Because of the use of multiple trees, compared to the decision tree, this algorithm reduces the probability of stumbling, which makes the prediction more credible. Besides, by creating multiple estimators, the influence of over-fitting is reduced.

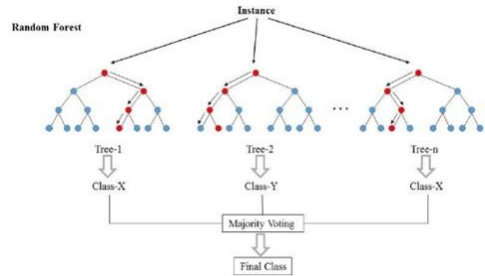

Figure 2: Random Forest Process

**AdaBoost** In AdaBoost, a meta-estimator begins by fitting a classifier to the original dataset. Subsequently, it fits additional classifier instances to the same dataset, with weights adjusted for incorrectly classified instances, so subsequent classifiers focus more on arduous instances. Adaptive boosting involves assigning different weights to each instance, with higher weights given to incorrectly classified instances.

**Model Stacking** Stacking is an ensemble machine learning algorithm that learns how best to combine the predictions from multiple well-performing machine learning models.

- Level-0 Models (Base-Models):Models fit the training data and compiled predictions
- Level-1 Model (Meta-Model):Model that learns how to combine the base models' predictions best

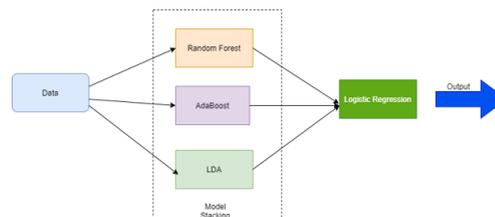

Figure 3: Model Stacking Framework used in this study

**Neural Network** Neural Network (NN) is a mathematical model of how the brain functions. By adding more 'Hidden Layers,' you can increase the number of layers. In the first layer, external information is received, corresponding to independent variables in statistics. Neurons in the input layer send signals to the hidden layer, and all information collected from neurons in the input layer is transmitted to hidden layers. Many financial prediction studies have used NNs since the 1990s, and most of these studies report that NNs have higher accuracy than conventional statistical techniques like LDA, QDA, Logistic regression, etc.

Credit scoring is performed using a multilayer neural network architecture, one of the deep learning architectures. Input neurons transmit signals to the hidden layer, receiving information from the input neurons. Our proposed algorithm produces promising results compared to the recent best performing papers that use P2P lending datasets.

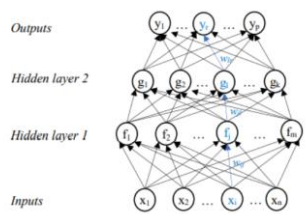

Figure 4: Two hidden layers neural network. The n inputs, m and k nodes of each hidden layer and p outputs.

**Hyper Parameter Tuning for Models** In this study, we trained several models by grid search on a balanced sampling from 2016 accepted loans data.
The models are:

- Decision Tree: $max\_depth=6$
- Random Forest: $max\_depth=10$ and $nestimators=100$
- Logistic regression: L2 regularization

- Adaboost: estimator=DecisionTreeClassifier (max_depth=6), learning rate=0.5 and n estimator=50
- LDA with default settings
- QDA with default settings
- Stacking: Mentioned above in description
- Neural Network: NodesperLayer:200,100,40,1('output' layer) (using Keras Sequential API)
  Activation function: ReLU
  Loss function: binary cross entropy Output unit: Sigmoid
  Optimizer: Adamax (use default settings) Epochs:100
  Batchsize:100
  Validationsize:0.3
  Early-Stopping is applied to mitigate overfitting

## V. Model Assessment

Measuring prediction performance is essential to evaluating machine learning performance. An error rate is commonly used as a measure of binary model performance. There are four different combinations of predicted and actual values in the confusion matrix, which helps to comprehend classification errors. The dataset records are arranged in a matrix according to the actual category and classification model prediction category. This matrix shows the actual value in the row and the predicted value in the column.

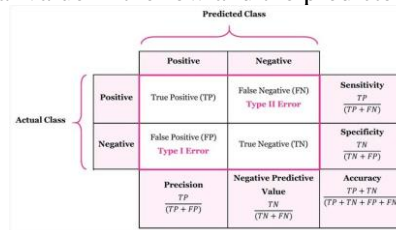

Figure 5: Confusion Matrix

| Name | Explanation |
| --- | --- |
| Accuracy (ACC) | Calculated as (TP+TN)/(TP+TN+FP+FN) |
| Precision (PPV) | Calculated as TP/(TP+FP) |
| Sensitivity (TPR) | Calculated as TP/(TP+FN) |
| Specificity (TNR) | Calculated as TN/(TN+FP) |
| FPR | Calculated as 100%−TN/(TN+FP) |
| ROC curve | Receiver operating characteristics curve (x-axis: FPR versus y-axis: TPR) |
| F1-score | 2xPrecisionxRecall /(Precision + Recall) |
| Log Loss | Probabilistic confidence of accuracy |

Table 1: Performance Indicator

### 5.1 Model Comparison

We have selected several performance indicators to evaluate various machine learning models to test the accuracy, precision, sensitivity, ROC-AUC score, F1 score, and Log- loss. One of the most critical evaluation indicators for machine learning models is the AUC-ROC curve.

| Model | Sensitivity | Specificity | Accuracy | AUC | F1 | loss |
| --- | --- | --- | --- | --- | --- | --- |
| Decision Tree | 0.667204 | 0.659842 | 0.663429 | 0.732591 | 0.658864 | 0.59046 |
| Random Forest | 0.720894 | 0.731023 | 0.725857 | 0.818094 | 0.728443 | 0.553895 |
| Logistic L2 | 0.677986 | 0.670666 | 0.674238 | 0.745122 | 0.670107 | 0.586086 |
| Adaboost | 0.788741 | 0.775363 | 0.781881 | 0.879406 | 0.778939 | 0.630922 |
| LDA | 0.679128 | 0.670357 | 0.674619 | 0.744691 | 0.669808 | 0.589595 |
| QDA | 0.954652 | 0.535525 | 0.565452 | 0.568231 | 0.238812 | 14.870424 |
| stacking | 0.670971 | 0.68221 | 0.676429 | 0.75165 | 0.680866 | 0.584733 |
| Neural Network | 0.776022 | 0.689021 | 0.724286 | 0.815701 | 0.695279 | 0.521027 |

Table 2: In-Sample Metrics

The ROC curves show an equal balance between true and false positives. A 2D plot of the performance of binary classifiers under threshold options such as false acceptance rate (FAR) and false rejection rate (FRR). The ROC curve encloses an area under the curve (AUC). The AUC is generally greater than 0.5 when the ROC curve plots above function y=x. If AUC=1, the classifier has perfect predicting power, and each sample's true value could be predicted correctly. If $0.5 < AUC < 1$, the classifier has certain predicting power under threshold settings.

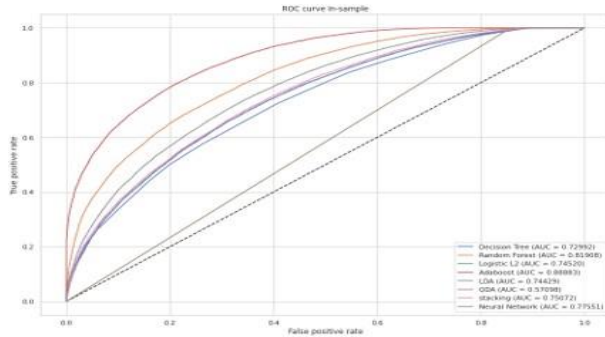

Figure 6: In-Sample ROC curve

| Model | Sensitivity | Specificity | Accuracy | AUC | F1 | loss |
|---|---|---|---|---|---|---|
| Decision Tree | 0.656147 | 0.646211 | 0.651056 | 0.720851 | 0.647115 | 0.606675 |
| Random Forest | 0.667975 | 0.672105 | 0.67 | 0.742813 | 0.673555 | 0.59878 |
| Logistic L2 | 0.677475 | 0.664689 | 0.670889 | 0.742572 | 0.666254 | 0.58667 |
| Adaboost | 0.661531 | 0.648678 | 0.654889 | 0.712583 | 0.649436 | 0.663227 |
| LDA | 0.679472 | 0.66339 | 0.671111 | 0.741951 | 0.664855 | 0.590636 |
| QDA | 0.949962 | 0.533681 | 0.563722 | 0.567879 | 0.239124 | 14.926757 |
| stacking | 0.667357 | 0.671501 | 0.669389 | 0.742541 | 0.672968 | 0.592187 |
| Neural Network | 0.699216 | 0.632957 | 0.659722 | 0.733931 | 0.624072 | 0.599666 |

Table3: Out-Sample Metrics

We compared the performances of Logistic regression, CART, Random Forest, AdaBoost, LDA, QDA, Stacking, and Neural Networks. The results from all the coefficients are shown in Table 2,3, and Figure 6, 7. Random Forest, Adaboost, Neural Network per- form better for In-sample Dataset. QDA has the lowest AUC but the highest Sensitivity as per Table 2.

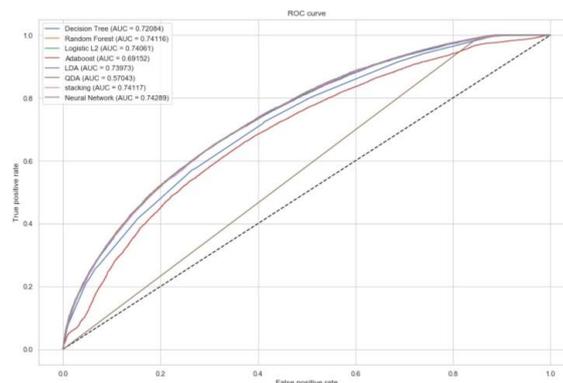

Figure 7: Out-Sample ROC curve

However, if we see for out-sample dataset result, we can Random Forest, Logistic Regression, LDA have close AUC=0.74 and Neural Network=0.73. We cannot distinguish the model with very high accuracy and AUC score for the test dataset.

### 5.2   Using Stratified K-Fold Cross-Validation

An investigation of how successful a model is at forecasting or anticipating an outcome is known as a modeling evaluation. The performance of a model must be estimated using data that already has the aim or outcome. Model evaluation includes experimenting with alternative data preparation strategies, learning algorithms, and hyperparameters.

- Model=Data Preparation + Learning Algorithm + Hyper-parameters

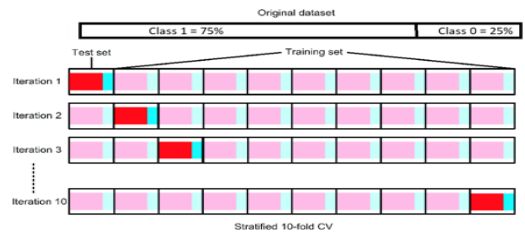

Figure 8: Stratified10-foldCV

Ideally, the model construction procedure (data preparation, learning algorithm, and hyper-parameters) with the best score (with your chosen metric) can be selected and used.

To evaluate a model, divide a dataset into two parts, train one, and test the other. Consequently, each part of the dataset has been named according to its function, trainset, and test set. If our dataset is representative and significant, this method works well. Depending on the problem, we may have to provide thousands, hundreds of thousands, or even millions of examples to satisfy the requirements. In this study, we have done a 70/30 split for the train and test we have considered. The study shows that there is a class imbalance in which

class1(Fully Paid) makes up almost 75%of the sample, and class 0 (Default) represents only 25%. Due to this, we can also observe overfitting and biases in in-sample results even though we do random splits and shuffles in train/test splits.

| Model | accuracy(mean) | F1 | AUC | std(mean) |
|---|---|---|---|---|
| Neural Network | 0.740131 | 0.709852 | 0.812901 | 0.00155 |
| LDA | 0.672517 | 0.67817 | 0.74261 | 0.001763 |
| Logistic L2 | 0.672317 | 0.678131 | 0.743449 | 0.002417 |
| Random Forest | 0.67125 | 0.669005 | 0.743578 | 0.00255 |
| stacking | 0.671017 | 0.668802 | 0.743485 | 0.002309 |
| AdaBoost | 0.657583 | 0.6585 | 0.7158 | 0.001781 |
| Decision Tree | 0.6563 | 0.6615 | 0.7229 | 0.002282 |
| QDA | 0.5649 | 0.694305 | 0.688129 | 0.0012736 |

Table 4: Performance Indicator using stratified K-Fold CV

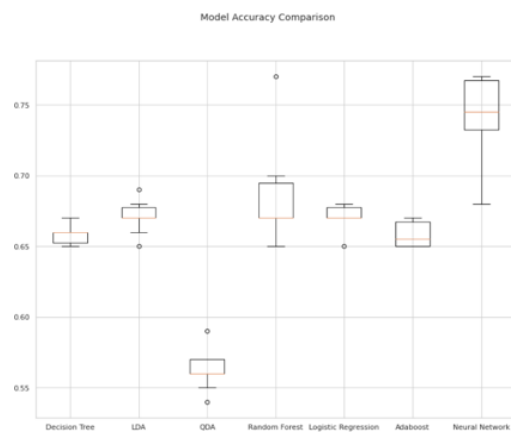

Figure 9: Model Accuracy Comparison Using Stratified10-fold CV

The10-foldcross-validation process is the most widely used model evaluation scheme for classifiers. When utilizing k-fold cross-validation or a train-test split, the answer is not to split the data arbitrarily when there is an imbalanced classification. We can, for example, randomly divide a dataset while preserving the same distribution of classes in each subset. The target variable (y), the class, is utilized to influence the sampling process, which is known as stratification or stratified sampling. A k-fold cross-validation variant can preserve the uneven class distribution in each fold. The stratified k-fold cross-validation ensures that the distribution of classes in each data split is the same as the overall training dataset distribution. After that, doing stratified K-fold cross-validation gave a means to improve a machine learning model's predicted performance. Table 4 and Figure 9 show that utilizing stratified10-foldcross-validation has improved the results, and Neural Network outperforms, and we can easily select the best model out of all.

## VI. Explainability Using SHAP & LIME

As a general rule, Machine Learning (ML) algorithms capable of capturing structural non- linearities in training data-sometimes referred to as "black boxes" (e.g., Random Forests, Deep Neural Networks) - are far better at predicting than their linear counterparts (e.g., Generalized Linear Models). However, they are somewhat harder to interpret quite often; given an instance of input data (i.e., the model features), it is not always possible to understand why a certain prediction has been made. Thus, it has not been possible to use ML algorithms that are "black boxes" in situations when clients are seeking cause and effect explanations for model predictions, rather, sub-optimal predictive models have been used instead, as their explanatory power is more valuable, in comparison. It is hard to define a model's decision boundary in a way that is understandable to humans. Shapely Additive Explanations (SHAP) provide a unified methodology for explaining the output of machine learning models. In this paper, we have interpreted our Neural Network model with SHAP and LIME.

### 6.1 LIME

"Locally Interpretable Model Agnostic Explanations is a post-hoc model-agnostic explanation technique that aims to approximate any black-box machine learning model with a local, interpretable model to explain each prediction", Ribeiro et al (2016). According to the authors, LIME can explain any classifier, no matter what algorithm is employed for predictions since it is independent of the original classifier. The LIME System works locally; this means it is observation-specific, and as with SHAP, it will provide explanations for every specific observation. "LIME works because it fits a local model based on similar data points to the case under consideration. Local models belong to the class of potentially interpretable models, such as linear models, decision trees, etc.", Ribeiro et al(2016) for more information. The explanations provided by LIME for each observation x is obtained as follows:

$$\xi(x) = \mathrm{argmin}_{g \epsilon G}\, L(f, g, \pi x) + \Omega(g(1)) \qquad (9)$$

where G is the class of potentially interpretable models such as linear models and decision trees,

$g \epsilon G$ : An explanation considered as a model
$f: R^d \rightarrow R$: The main classifier being explained
$\pi x(z)$ : Proximity measure of an instance z from x
$\Omega(g)$ : A measure of complexity of the explanation $g \epsilon G$

A core feature of LIME is that it does not assume anything about f (since its primary criterion is model agnostic). L is the measure of the unfaithfulness of g ś approximation of f in the locality defined by $\pi(x)$.

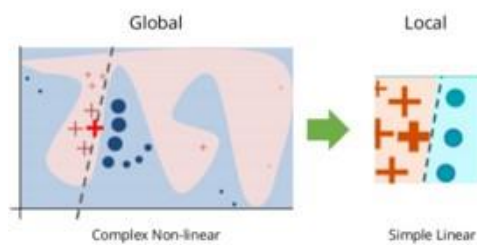

Figure10: Source: "Why Should I Trust You?" Ribeiro et al (2016)

#### 6.1.1 Using LIME to explain Local Instances

As part of our specific use case, we utilized the LIME package available in the Python library. To obtain the model explanations, we followed the following steps:
- Create a list of all the features' names
- Determine whether the target labels are "Fully Paid" or "Default"

- A function that produces class probabilities will be fed an array of test cases (feature values)
- Pass the training data, feature list, class list, and probability function to the lime explainer object
- Using the explainer, select an instance and send it as a parameter to the explainer.

The main features that have contributed to pushing the probability towards either "Fully Paid" or "Default" are shown in Figure 4. An output is then displayed with a list of the top 10 features to the model's prediction.

### 6.1.2 LIME on Neural Network Based Models

We attempted to apply the approach to the trained Neural Network classifier after identifying the processes involved in collecting the explanations from the LIME implementation. The primary issue we ran into in this situation was creating the LIME framework in Python, which is incompatible with models that use GridSearch CV objects. As a result, we retrain the model with the best parameter set found during the hyperparameters grid search, and the model is saved as a Neural Network object.

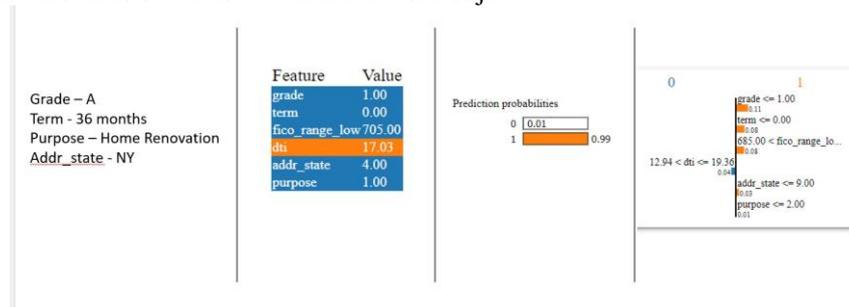

Figure 11: LIME explanation for a customer classified as a "Fully Paid" Loan type by Neural Network Model

We picked another use case to explain how Lending Club assigns grades to different loans.

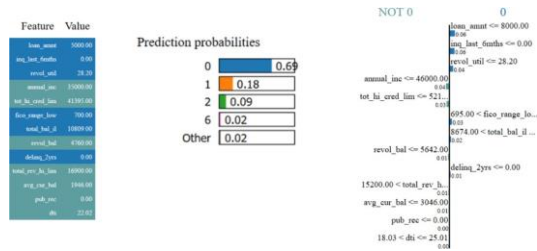

Figure12: LIME explanation for a Loan classified with different grades (0– 'A',1-'B', 2-'C',3-'D',4-'E',5-'F',6-'H',Others-any) Loan type by Neural Network Model

### 6.1.3 Interpretability

To expand on the explanations provided by the LIME framework, we interpret three examples in this section. Figure 11 shows an example of a loan for which the model predicts a class "1" outcome (i.e., the loan contract will not default). Further, Figure 11shows the top six factors that contributed to this decision and their contributions. On the left is the model's confidence in its prediction. We can interpret this as follows:

- Since the value of the "dti" variablewas17.03, which is between12.94 and19.36(a value used by the model for making a decision), this pushed the prediction towards the "Charged off" category
- Similarly, we see categorical features such as grade, term, addrstate, purpose, these model pushes the prediction towards "Fully Paid"

In general, the model is confident that the loan contract will not default, given the customer's feature values. The LIME framework can also allow model developers to explain to end-users the reasoning behind a particular decision. In Figure 12, we present our second example of a loan contract for which the model has predicted different Lending Club grades (i.e., Grades that LC assigns to each loan). Figure 12, similar to Figure 4, illustrates the top thirteen features that contributed to this decision along with their respective contributions. On the left is the model's confidence in its prediction.

**6.2    SHAP**

According to the paper by Lundberg et al. (2017), "Shapley Additive exPlanations(SHAP) presents a unified framework for interpreting predictions. "SHAP assigns each feature an importance value for a particular prediction". Its novel components include:
- The identification of a new class of additive feature importance measures
- Theoretical results show there is a unique solution in this class with a set of desirable properties

$$g(z') = \phi_0 + \sum_{j=1}^{M} \phi_j z'_j$$

(10)

Where g again, is the explanation model, $z' \epsilon 0,1^M$ is the simplified feature, M is the maximum coalition size and $\phi_j \epsilon R$ is the feature attribution for a feature j, the Shapley Values. In the simplified features, an entry of 1 means that the matching feature value is "present" and 0 that it is "absent". To compute Shapley Values, SHAP simulates that only some of the feature values are "present" and some are "absent". This is equivalent to playing or not playing in the coalition from a game theory perspective. SHAP represents the coalitions a linear model to compute the $\phi's$. Referring back to our instance x the simplified features $x'$ is a vector of all features that are "present" (Lundberg et al.(2017).

The goal is to determine the impact each feature has on the target. The SHAP value corresponds to the change in the model prediction based on that feature. They explain how to get E[f(z)] to the current output f(x) from the predicted value used if the features are unknown. The paper does not go into the technical details

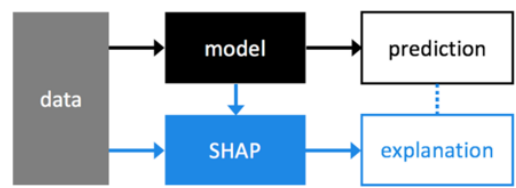

Figure13: SHAP process followed in the paper

SHAP, which are beyond the scope of the paper. See Lundberg et al. [5] for more information on SHAP and the theorems developed by the authors.

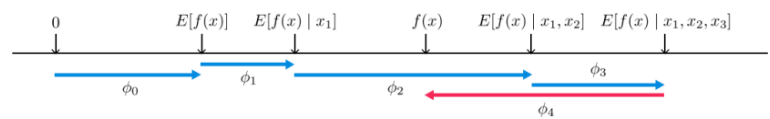

Figure 14:"SHAP (Shapley Additive exPlanation) values attribute to each feature the change in the expected model prediction when conditioning on that feature", (Lundberg Lee, (2017)

**6.2.1    Model Development Using SHAP on Neural Network**

The NN classifier was developed through a series of trials. We adjusted the number of hidden layers and the number of neurons in each layer to create different model topologies. During this experiment, we switched between "sigmoid" and "relu" learning rates and activation functions. Because we were working with binary classifications, we employed the binary cross-entropy loss function. After several trials, the model with the greatest ROC AUC score on the test set was chosen. Table 2 contains the final model parameters. 5.2.2 We utilized the deep explainer and the kernel explainer to explain the Neural Network classifier. As previously said, we experimented with 50,100,1,000, and 2,000 test data samples and reported the results of our findings. When we first started with the deep explainer, we observed significant variations between the tree and kernel explainers. When comparing Deep explanations to tree explanations, the SHAP values for both classes were

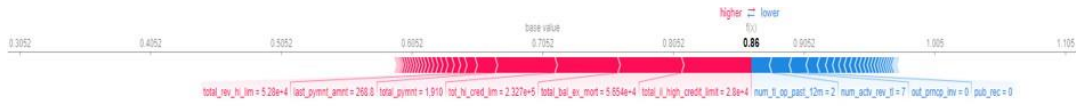

Figure15: Shapley Force Plot for a high score observation (for Numeric features)

pooled in Deep explanations, whereas the SHAP values for each class were presented individually in tree explanations (i.e.,SHAP values are given individually for each class).Despite the fact that two principles are diametrically opposed, one argument might be used to explain the other. Deep explanation also has a faster computing time than kernel explainer. Because model specific explainers may support specific model designs, the evidence shown above shows that utilizing them leads to substantially faster computations. Individual prediction reasons can be examined using SHAP, as seen below. For instance, we have two observations with different attributes that result in high scores and cause the system to forecast a "Fully Paid" loan.

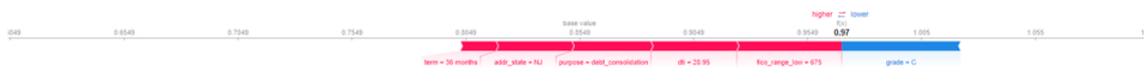

Figure16: Shapley Force Plot for a high score observation (for Categorical features)

Figure 17 shows the summary plot. It helps us overview which features are most important for our model for every feature and every sample. SHAP values are used to visualize the distribution of each feature's impacts on the output of the model (red high, blue low) using the sum of SHAP values over all samples. For example, term one year is blue, and term five years is indicated by red. In short than long term, it has a highly negative impact, and a Low Fico score is a high negative impact.

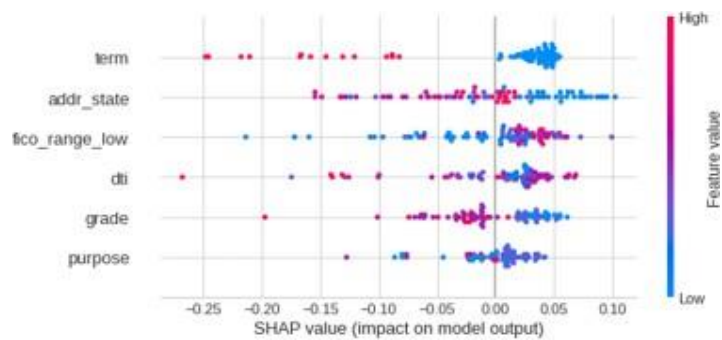

Figure 17: Neural Network model using deep Kernel Explainer

## VII. A Novel Optimal Threshold-Based Investment Strategy

Excellent performance on the test set does not necessarily translate into a significant Return on Investment (ROI). We want to determine the optimal cutoff in the predicted probability to maximize ROI using our optimized models. The threshold determines whether a projected probability or score is a class label. If the threshold is changed, the accuracy and profit values change. For normalized projected probabilities between

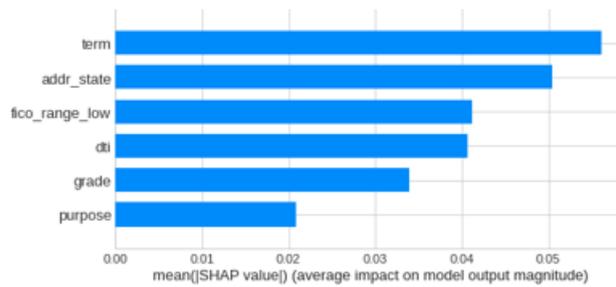

Figure18: Absolute impact on target variable (Loan status)

0 and 1, the threshold is set to 0.5 by default. Classification models predict a positive class score for each event. If an event's score exceeds 0.5, it is automatically assigned to the positive category; otherwise, it goes to the negative category. By changing the classification threshold, we change the classification of positives and negatives. These changes in classification will change the accuracy and profit expectations.

| | ROI | Threshold | sensitivity | specificity | accuracy | ROC_AUC | Accepted Rate | loss |
|---|---|---|---|---|---|---|---|---|
| Decision Tree | -1.319906 | 0.5 | 0.462739 | 0.818874 | 0.689334 | 0.730258 | 0.636263 | 0.589294 |
| Random Forest | -0.147265 | 0.5 | 0.464479 | 0.850181 | 0.686246 | 0.767240 | 0.574969 | 0.585223 |
| Logistic L2 | -0.437311 | 0.5 | 0.461146 | 0.837668 | 0.684915 | 0.749112 | 0.594305 | 0.574606 |
| Adaboost | 0.743650 | 0.5 | 0.470952 | 0.850573 | 0.692204 | 0.759345 | 0.582824 | 0.658383 |
| LDA | -0.416361 | 0.5 | 0.461236 | 0.837241 | 0.685064 | 0.748173 | 0.595278 | 0.578509 |
| QDA | -5.190200 | 0.5 | 0.870726 | 0.744590 | 0.750360 | 0.570982 | 0.954253 | 8.500340 |
| stacking | -0.363417 | 0.5 | 0.450754 | 0.842773 | 0.674050 | 0.750267 | 0.569606 | 0.582367 |
| Neural Network | -0.281208 | 0.5 | 0.470943 | 0.838874 | 0.693475 | 0.757035 | 0.604821 | 0.573726 |

Table5: Performance Indicator w.r.t ROI when threshold is 0.5

Our optimization of an economic target function is based on the above-explained algorithms. To determine the maximum threshold value to determine default assets and non-default assets where we are getting maximum profit, we played with the threshold limit and compared the profit function for each threshold.

ROI is calculated as the net return from investment over the amount of input. A naive way of formalizing it would be to divide net gain by investment amount. In this approach, we only consider the return-on investment rate from the invested loans. Hence, we determined the threshold for each model to maximize naive ROI. We then use the grid search method to find the best prediction threshold for each model to get maximum profit.

- The formula of ROI is given by:

$$ROI = \frac{(Gain from Investment - Cost of Investment)}{Cost of Investment}$$

(11)

The nominator (Gain from Investment- Cost of Investment) is also defined as the net return on investment (NRI). Net return on investment (NRI) is a way to encode the profit of an investor, and the model is optimized for this economic target function.

As the main focus of an investor is not to maximize prediction accuracy but rather to maximize its profit, we added ROI diag vs. threshold to see the impact of threshold on ROI. So, we optimize not for accuracy but an economic target function. Here we could see clearly that threshold optimized via ROI is more conservative. For each table, we could see that all models except QDA tend to have high specificity but low sensitivity on actual data, where the majority of the accepted loans are fully paid.

| | ROI | Threshold | sensitivity | specificity | accuracy | ROC_AUC | Accepted Rate | loss |
|---|---|---|---|---|---|---|---|---|
| Decision Tree | 2.357002 | 0.82 | 0.307117 | 0.947600 | 0.366354 | 0.730258 | 0.092487 | 0.589294 |
| Random Forest | 3.677177 | 0.80 | 0.294385 | 0.976288 | 0.321659 | 0.767240 | 0.039997 | 0.585223 |
| Logistic L2 | 3.806862 | 0.89 | 0.290583 | 0.967195 | 0.309017 | 0.749112 | 0.027246 | 0.574606 |
| Adaboost | 3.027355 | 0.51 | 0.368263 | 0.906811 | 0.534098 | 0.759345 | 0.307930 | 0.658383 |
| LDA | 3.825114 | 0.89 | 0.289013 | 0.972176 | 0.303277 | 0.748173 | 0.020880 | 0.578509 |
| QDA | -5.180625 | 0.89 | 0.861651 | 0.744779 | 0.750241 | 0.570982 | 0.953270 | 8.500340 |
| stacking | 3.443606 | 0.89 | 0.289196 | 0.975729 | 0.303804 | 0.750267 | 0.021277 | 0.582367 |
| Neural Network | 3.834185 | 0.89 | 0.298739 | 0.965282 | 0.337062 | 0.757035 | 0.057495 | 0.573726 |

Table6: Performance Indicator for optimal *p*(threshold) value for various Machine Learning models to maximize profit

Comparing the number of defaulted loans to the number of non-defaulted loans, we immediately notice the small number of defaults. Finally, to classify loans, we need to introduce a threshold $p_{thr}$ on the predicted default probability to help decide which loans will be accepted and rejected. The threshold determines which loans are accepted and rejected based on predicted default probabilities. For example, a threshold of 1 would reject all loans, which would be the correct decision for all defaulting contracts but would be the incorrect decision for all non-defaulting ones, therefore giving the same(low) ratio of defaulting loans. An investor accepting defaulted loans at a threshold of $p_{thr}=0$ would suffer heavy losses.

The "Profit by threshold" concept involves assigning scores to applicants in the validation set (Gramespacher T et al. (2021)). Based on the credit score, there are two creditability classes: "risky" and "creditworthy" scores predicted by various models that we used above for credit risk management and a classification threshold. The classification is repeated multiple times, starting with a low threshold value and increasing it for each iteration. The output table contains the accuracy statistics, the number of accepted loans, and the expected ROI obtained using the different threshold values. Identifying and rejecting at least some of the defaulted loans has a huge positive influence on the investor's profits (Gramespacher T et al. (2021)), greatly outweighing the cost sustained by rejecting some of the non-defaulting loans. Despite the fact that rejecting good business does not appear appealing, the huge rise in the bank/investor's profit is a compelling justification for using a machine learning model. All of the above findings, including the considerable increase in profit, may be found not only in the training set (in-sample), but also in the test set (out-of-sample)—which has not been used in either the model training or the threshold optimization.

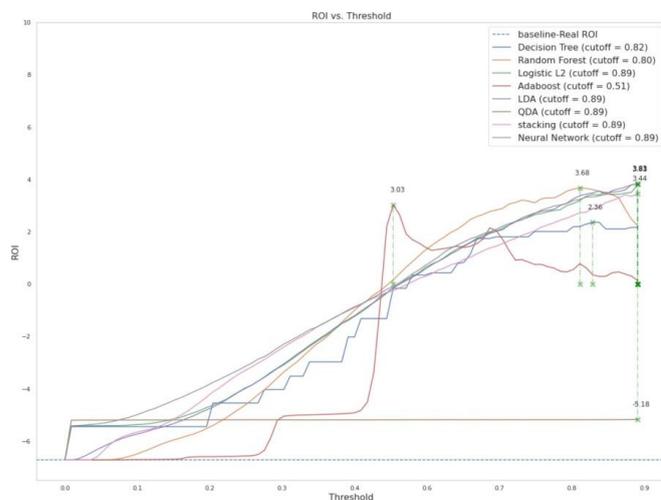

Figure19: ROI vs Threshold Graph for various ML Models

It is critical to consider the user's target function when selecting the optimal model and parameters to optimize profit. We modified the threshold distinguishing between charged off and fully paid loans in machine learning models to maximize the provided profit target function. We find that profit-maximizing models reject a considerable number of loans; that is, these models accept a large number of mistakenly rejected good firms to sort out a few more defaulting loans.

## VIII. Conclusion

Several machine learning models were analyzed and compared in this study using a variety of performance indicators. LIME and SHAP have been utilized to explain Black-Box classifiers such as Neural Network and we have also worked towards improving investor profit and focused on transparency for the selection for loans/assets by optimizing loan decisions. Investors may benefit from our research. Currently, most investors base their investment decisions solely on the loan grades. Selection of investment assets is both an art and a science. By identifying the significant elements in our study, we have created the groundwork for explainability in lending and loan investment selection with some level of transparency and automation using machine learning models.